# What Communication Modalities Do Users Prefer in Real Time HRI?

Ori Novanda[1,2,3], Maha Salem[3], Joe Saunders[3], Michael L. Walters[3], and Kerstin Dautenhahn[3]

**Abstract.** This paper investigates users' preferred interaction modalities when playing an imitation game with KASPAR, a small child-sized humanoid robot. The study involved 16 adult participants teaching the robot to mime a nursery rhyme via one of three interaction modalities in a real-time Human-Robot Interaction (HRI) experiment: voice, guiding touch and visual demonstration. The findings suggest that the users appeared to have no preference in terms of human effort for completing the task. However, there was a significant difference in human enjoyment preferences of input modality and a marginal difference in the robot's perceived ability to imitate.

## 1 INTRODUCTION

Humans often use multi-modal interaction in daily communication and frequently use speech, physical gesture, and eye gaze when communicating with each other. In contrast, people do not usually interact with machines in the same way they interact with other humans. For example, when we open the fridge door in the morning, we do not usually greet it as we would another person.

With the recent advances in technology, it is now quite common for people to speak to some machines. High-end consumer products such as smartphones and tablets have enough computing power to capture human speech and translate it into text commands. This allows people to use their voice to interact with the applications running on the device. This technology has given rise to digital virtual assistants such as: Siri [1] on the iOS platform, Google Now [2] on the Android platform, and Cortana [3] on the Windows platform. These systems enable people to get information simply by asking the device. For example, asking what the weather will be like, or when a flight will leave. Language learning programs, such as Duolingo [4], prompt users to say sentences and use a voice to text translation method to accept their answer.

Traditionally robots have been associated with factories for building products such as cars. However, robots are now increasingly being used in a number of application areas where people can interact with them in a more natural way, in some ways similar to how they would interact with living creatures, such as indicated in the survey by Leite et al. [5]. For example, Pleo [6] changes its behaviour depending on how the user interacts with it, and Fernaeus et al. [7] used it to learn how people play with a robotic animal. KASPAR, a child-size humanoid robot, has primarily been developed as a mediator to interact with children with autism in order to encourage basic communication and social interaction skills [8]. The consumer and research robot NAO [9] has been programmed to fulfil many tasks, one of which is as a companion robot (see Dautenhahn [10]) such as used in the research by Baxter et al. [11].

Since Sheridan [12] first associated Human-Robot Interaction (HRI) with teleoperation of factory robotic platforms, HRI research has extended into a number of different research areas (Goodrich and Schultz [13]). One of the areas of particular interest in recent years is multi-modal interfaces for multi-modal interactions. Stiefelhagen et al. [14] suggested that multi-modal interfaces are required to facilitate natural interaction. When humans are interacting with machines that have some human-like characteristics, they have a tendency to anthropomorphise with the machine and communicate in ways similar to human-human communication [15]. One of the objectives of HRI is to make human-robot interaction easier, more intuitive and more user friendly. By providing a multi-modal interface it may help keep the users engaged and interact with them in a more familiar manner, similar in some ways to which they may interact with other humans.

Although interactive multi-modal systems have some distinct advantages, developing such systems poses many challenges. According to Turk [16], the performance of a multi-modal system depends on each unimodal technology. Currently each modality has its own ongoing progress as an active research field. For example, a survey by Argall and Billard [17] lists research that solely focuses on investigating the tactile input modality.

Developing multi-modal interactive systems requires a substantial amount of computing power and robust integration algorithms. The integration algorithm of the robot's sensing system needs to make decisions in real-time on which input to consider for giving an appropriate response or action through the robot's actuators. The system has to be powerful enough to process different inputs such as visual, audio, and gesture cues. Integrating these social queues to flow naturally throughout the interaction session will also consume additional processing power. Providing a robust input modality and fusion to integrate all input data is a technically challenging task. Many hours of work would need to be devoted just to prepare the robot for a relatively simple task. This is one of the reasons that some HRI studies use Wizard-of-Oz [18] approaches to run experiments. By using these approaches, limitations on the technology can be set aside and replaced by behind-the-scene controllers to produce behaviour for the robot which is perceived by users as autonomous.

The challenge of creating a multi-modal interactive robotic system has inspired the research in the current study which investigates users' preferences of input modality when providing information to a robot. The study was designed to ask users to experience three different modalities whilst delivering the same instructions to the robot.

[1] Dept. of Electrical Engineering, Universitas Sumatera Utara, Indonesia, ori@usu.ac.id
[2] This author received a scholarship from the General Directorate of Higher Education of Ministry of Education and Culture of Indonesia
[3] Adaptive Systems Research Group, University of Hertfordshire, United Kingdom

## 2 RELATED WORK

The study took related research in Human-Computer Interaction (HCI) into consideration. As suggested by Kiesler and Hinds [19], and Breazeal [20], existing work in HCI offers rich resources and inspiration for research in HRI.

The experiment "Put That There" by Bolt [21] is widely considered a pioneering demonstration that first showed the value and opportunity of multi-modal interfaces over uni-modal interfaces in HCI. The experiment was conducted using speech and gesture as command channels to draw a map.

The multi-modal interface raised a question of when the system is capable of multi-modal interactions, will the users utilise the ability to interact multi-modally? Oviatt [22] discussed ten myths about multi-modal interaction that give useful guidance to researchers building multi-modal systems. He stated that with multi-modally capable systems, users tend to switch between uni-modal and multi-modal interaction with the multi-modal interactions being the most predictable, based on the type of action being performed. In a previous study Oviatt et al. [23] found that 86% of the time participants used multi-modal commands when navigating a map in order to move, add, modify, or calculate the distance between objects. For performing tasks that require no navigation of the map, such as printing the map, the participants interacted multi-modally less than 1% of the time.

Later, Oviatt et al. [24] conducted an experiment using a Wizard-of-Oz approach, and concluded that the cognitive load of the task will drive the users' preference towards either uni-modal or multi-modal interaction. Tasks with higher difficulty will often cause the users to utilize the multi-modality of the system. With repetitive tasks, users would initially communicate multi-modally. Once the tasks became more familiar they then tended to prefer one particular interaction modality four times more often than interacting multi-modally.

Schüssel et al. [25] experimented using speech, gesture, and touch in multi-modal interactions to select graphical icons on a computer monitor. This experiment was also conducted using the Wizard-of-Oz approach and measured what modality was used and combined by the users to complete the task. The overall results of the modalities used were: touch (63.2%), speech (21.6%), gesture (11.2%), speech+gesture (3.6%), speech+touch (0.5%). None of the participants used speech+gesture+touch at the same time.

Carbini et al. [26] observed users' preferences for using a story telling game. Each user was given a task to compose a coherent story from a set of objects on a computer screen. It was found that children could easily interact using speech and gesture as compared to adults. The results of the full dataset were: gesture (45%), speech (5%), gesture+speech (50%).

All of the research cited above was conducted in HCI domains, where the users interacted with computers. This current research is focused on the interaction between humans and robots. Presented below are some studies that are more closely related to research in HRI.

Research by Khan [27] surveyed 134 respondents about their preferred interaction modalities with a robot. One of the questions asked in this survey was the preferred method of communicating with a service robot to take care of clothes on a couch, or when the robot is to inform the user that the task has been completed. The results showed that speech was the most preferred interaction modality (82%), followed by touch screen (63%), gestures (51%), and typing commands (45%). However, the results of this study are limited because the survey was conducted by asking participants to complete a questionnaire without the participants having interacted with an actual robot.

Salem et al. [28] conducted research to compare the preference of modality in HRI. In contrast to the current research, they investigated the output side of the multi-modal interface. They examined the perceptions of users regarding a robot when the robot provides information to the human uni-modally (voice only) and multi-modally (voice and gesture). It was found that the robot was evaluated more positively if it displayed non-verbal behaviours, such as hand and arm gestures along with speech, even if they do not semantically match the spoken utterances.

Humphrey and Adams [29] also conducted a study relevant to our current research, by measuring users' preference for visualising a tele-operated robot's compass. They compared two different compass visualisations: top-down and world-aligned. The top-down visualisation received higher preference, but there was no significant difference to the world-aligned visualisation

## 3 THE STUDY

The study presented in this paper builds on two main observations from the related work discussed above which are:

1. As described in [24], simple task interaction can be conducted sufficiently using a uni-modal system only.

2. Previous research established significant differences of modality preference one over another and the most-preferred modality also differed ([25], [26], and [27]).

Those considerations above come from the HCI research domain where humans interact with computers. This study puts them in HRI perspective, where humans interact with robots, to see whether they can be applicable to the HRI domain.

Based on the first observation (1), our research investigated further the modality comparison by conducting an experiment that asked users to do a simple-task, comparing the using of specific and different modalities in different sessions. Based on the second consideration (2), the study also evaluated which modality was most preferred.

This research aimed toward developing an autonomous humanoid robot that can perform a real-time multi-modal interaction. The developed system provides the capability to detect voice commands, and interprets gestures and touch. All processes run in parallel in real-time. In the discussion section, this paper presents the comparison of user preferences for the three input channel modalities when instructing the robot to move its arms.

The basic idea of the experiment for the research was to develop a robot that can be taught to dance following music. This idea was limited in the required capability in order to match the robot's physical limitations in speed of movement. The dance was changed to a simple mime task, and the music was limited to a single nursery rhyme. With these changes, the experiment became teaching the robot to mime following a nursery rhyme. The robot could be instructed to move its arms using voice commands, by the users' gestures, and by physically guiding the arms.

The experiment was run non-intrusively so that the users did not need to use gloves or markers. The users also did not have to wear a microphone or headphone. The voice command system used a speaker-independent system so it did not have to be trained prior to the experiment.

## 4 EXPERIMENT SETUP

This section describes the experimental setup for the study. The study was approved by the University of Hertfordshire Ethics Committee under protocol number a1213/10.

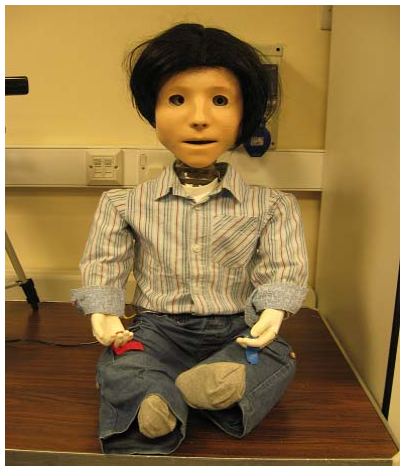

**Figure 1.** KASPAR Robot

### 4.1 The Robot

This research uses KASPAR [30], a child-alike humanoid robot (shown in Figure 1). It has 17 Degrees of Freedom (DoFs) and has an internal PC to run the robot autonomously. The robot uses eSpeak [31] text-to-speech engine for speaking.

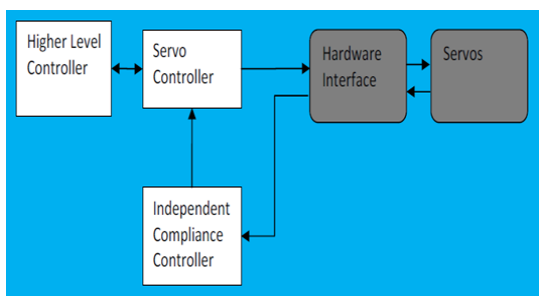

**Figure 2.** Compliance Mechanism

For the study, a program was developed to feature a servo compliance system. The block diagram of the compliance system is shown in Figure 2. It has a controller that measures the servos' torque values. This measurement is used to allow the software to detect whether the arms are being moved by an external force. It will then adjust the servos' positions to comply with the external force. With this feature, users can move KASPAR's arms without breaking the servos. This controller works independently and can override any arm movement commands sent by the higher level controller.

In the current implementation, there was a time delay in the compliance controller's loop path introduced by the hardware interface. This made the control bandwidth of the servos only achieve 1 Hz, which is lower than the human force control bandwidth which is around 20 Hz [32]. This made the arms slightly stiff to move.

The system used an additional external PC beside the internal PC. The PC's communicated using TCP/IP through an Ethernet connection. The robot was built to have a WiFi connection as well but this wireless connection was never used in the experiment because of the latency in data transmission.

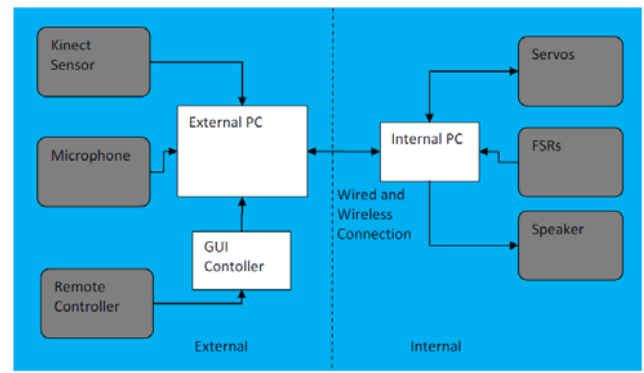

**Figure 3.** System Architecture

The external PC runs the high demand processes, such as the gesture detection and speech recognition. The global architecture of the system can be seen in Figure 3. The GUI controller runs on the external PC and sends commands to the internal PC to control the robot. The robot has several force sensitive resistor (FSR) sensors to detect touches. They are located on both palms and on the upper arms. This research did not restrict the participants on where they could touch the robot when moving its arms. During the experiment, the system only used the compliance system mentioned above to allow the participants to move the robot's arms physically.

### 4.2 Sensors

KASPAR was equipped with sensors to provide the following input modalities: (i) voice command, (ii) gesture, and (iii) touch. The developed system uses the Microsoft speech recognition engine. With non-intrusive interaction in mind, the system uses a directional microphone to listen to the user's voice. The microphone location was adjusted so the sound coming from the robot (voice and mechanical servo movements) was less likely to disturb the user's voice.

The speech recognition engine was programmed to detect 5 different commands that could be used to instruct the robot to move its arms. The robot has colour markers on its fingers (see Figure 1) to refer to the arms by colour instead of left and right (the former was deemed to be easier for participants to use when facing the robot). The markers are red and blue. The commands are: (i) red up, (ii) blue up, (iii) arms open, (iv) red down, and (v) blue down. As suggested by the name, 'up' and 'down' commands will instruct the corresponding red or blue arm to go up or down. The 'arms open' command will make both arms open wide.

The system could only detect one particular command at a time. After saying a command, the user was expected to wait for the robot to respond before saying the next command.

A Microsoft Kinect was used by the system to detect the human partner's gestures. The Kinect SDK provided a skeleton representation of the user's position and pose. The position of the wrists were measured and interpreted as commands to move the robot arms. The system was programmed so that it only detected 5 positions, which were equivalent to the 5 voice commands.

Touch input modality was provided to the robot by using the developed compliance system. The users could move the robot's arms by moving the arm directly. They could hold any part of the arm in order to move it e.g. the users could move the arms by moving the upper arm or moving the hand. The latter requires smaller force because it is further away from the shoulder joint.

### 4.3 Layout

The physical layout of the experiment is shown in Figure 4. The robot was 'sitting' on the table and the Kinect sensor was located next to the robot. Video cameras were used to record the activities during the experiment sessions.

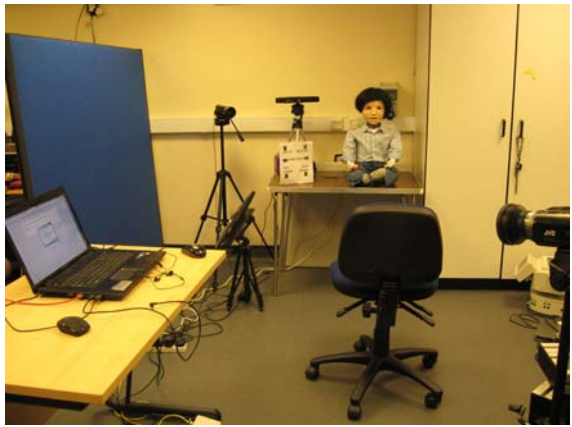

**Figure 4.** Experiment layout

Next to the robot was an instruction sign (see Figure 5) which reminded the user of the five instructions that could be used to control the robot. The instruction sign showed arrows to reflect the direction of the arms movement.

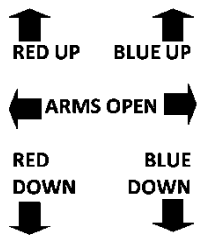

**Figure 5.** Instruction sign

### 4.4 Interaction Scenario

The task given to the participants in this study was teaching a humanoid robot to mime to a rhyme. The rhyme was 'Hickory Dickory Dock'. The participants had to instruct the robot to move the arms to mime by following the lines of the rhyme. The task was repeated in several sub-sessions by only allowing one or two of these modalities in each session: voice, gesture, touch, and voice+gesture.

### 4.5 Experiment Procedure

Before starting the experiment, the participants completed a demographic questionnaire and signed a consent form.

The experiment was divided into two main sessions:
1. Introduction session

In the beginning, the participant was introduced to the robot and asked to shake its hand. This was to familiarise the participants with the robot, and to let them know that it was fine to physically move its 'red arm' (right arm), even though it felt slightly stiff. Next, they were introduced to the nursery rhyme, and told what to do during the main trial session. The participants were also instructed on how to move the arms using each input modality.

During the introduction session, the robot was operated semi-autonomously using a wireless clicker to advance between sub-sessions. At the end of the introduction session, the participants were told that the following was the main trial, and the robot would run fully autonomously.

2. Main trial session

In the main trial, the participants were left alone interacting with the robot which ran autonomously. The investigator stayed in the same room reading a book and sat back-facing the participants at a table without any computer or electronics devices. The participants were told that in case of emergency or if they wanted to stop, they could notify the investigator at any time.

The trial was run individually with a single participant for each trial session. The robot first asked the participants to instruct it on how to move in order to follow the nursery rhyme. The robot said the rhyme, and the participant should then instruct the robot to move for each line of the rhyme. The participant could instruct the robot to move the arms while the robot said the rhyme, except in the voice command mode session, where the participants were instructed (by the robot) to say the command after the robot has finished saying the rhyme. In the touch modality sessions, the participants had to move forward close to the robot to move its arms.

In total, there were 4 sub-sessions in the main trial. Each sub-session presented to the participant a different input modality. The first three were arranged so each participant had a different order of voice, gesture, and touch modalities. In total there were 9 possible different orders. In the fourth sub-session, the participant was asked to instruct the robot using a freely chosen combination of gesture and voice commands. After each sub-session, the robot performed the complete 'dance' with movements and timings specified by the commands that had been given by the participant.

After the main trial session, a second questionnaire recorded the users' preferences of the methods to teach the robot. Before the whole session ended, the participants were also asked verbally whether they had any comments they wanted to express regarding the experiment.

### 4.6 Dependent Measurements

The post-trial questionnaire asked four questions using the Likert scale, and the participants rated their answers on a scale from 1 to 5. The first one was "Did you fully understand what instructions KASPAR said during the main session?" (1 being "not very well" and 5 being "very well").

The second question was "In terms of effort, how did you feel about the different methods to teach KASPAR to dance?" (1 being "very hard" and 5 being "very easy").

The third question was "In terms of enjoyment, how did you feel about the different methods to teach KASPAR to dance?" (1 = least enjoyable, 5 = most enjoyable).

The fourth question asked "When KASPAR showed what it had learned, how well did you feel KASPAR followed your instruction?" (1 = not very well, 5 = very well).

Every question from 2 to 4 had separate answers for each interaction modality.

## 5 RESULTS

The experiment was conducted with 16 participants; six females and 10 males aged 20 to 48 years old. They were recruited from the university staff and students. The invitation was advertised verbally and they were given a link of an online scheduler (Doodle [33]) to pick the available time slots that were suitable for them. In each gender category, 1 person was very familiar with robotic systems, while none had a prior knowledge of the robot setup that was used in this experiment.

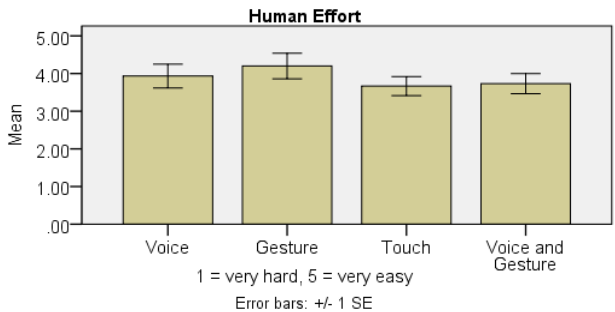

**Figure 6.** Questionnaire result on human effortlessness

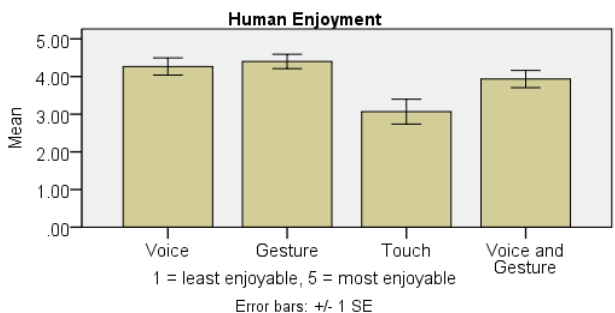

**Figure 7.** Questionnaire result on human enjoyment

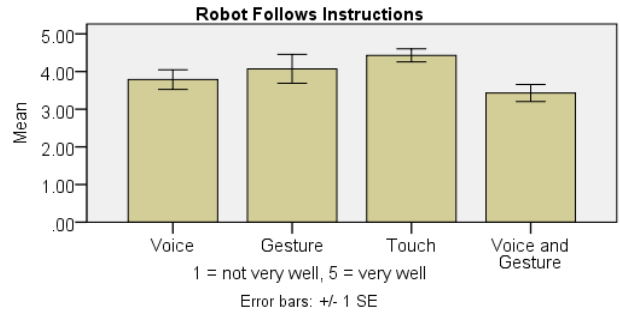

**Figure 8.** Questionnaire result on different instruction modalities

For the first question of the questionnaire, that asked whether the participants fully understood what the robot said during the experiment, no participant selected a value lower than 4. The mean score was 4.56 (SD = 0.51). The middle point of the answer was weighted as 3.

The questionnaire result on the effort to teach the robot to dance is shown in Figure 6. The data were checked using one-way repeated-measures ANOVA. The result was $F(3,42) = 0.848$, $p = 0.476$, which meant none was significant. The result suggests that no particular modality is perceived as harder than the others.

The result that is shown in Figure 7 shows participants' perceived enjoyment of conducting the task for each modality. The touch modality received the least enjoyable rating. The statistical analyses indicated a significant difference in preferences, $F(3,42) = 6.461$, $p = 0.001$. The pairwise comparisons results indicated that there was a significant difference ($p = 0.008$) between participants ratings for gesture (M = 4.4, SD = 0.74) and touch (M = 3.07, SD = 1,28) interaction modalities.

Finally, Figure 8 shows the participants' perception of the robot's ability to follow instructions. The difference was marginally significant, $F(3,39) = 2.56$, $p = 0.069$. The pairwise comparisons showed a preference ($p = 0.011$) for touch (M = 4.43, SD = 0.65) over voice+gesture (M = 3.43, SD = 0.85).

## 6 DISCUSSION

This research has investigated a robotic system that can be taught movements to follow a nursery rhyme. The development of the software is only presented briefly as it would be better to be presented as a technical paper. Three modalities were provided as input channels to give information to the robot as commands to move its arms. They are voice, gesture, and touch. Two modalities were provided as output channels: voice and gesture. The robot operated autonomously during individual sessions. The robot had touch-compliance which allows humans to physically move its arms into a desired pose. The system supported integration of multiple modalities through a TCP/IP-based inter-process communication mechanism. The experiment was conducted with adult participants.

The research findings indicated that being given a task which was to teach a robot to mime actions that follow a nursery rhyme, there was no statistically significant difference in preference ratings regarding human effort.

In contrast, there were favourable preferences regarding the human enjoyment. The touch modality was the least preferred and the gesture modality was rated the highest. The authors argue that the touch modality scored lowest due to the participants worrying about breaking the arms of the robot. This was because the compliance only controlled the arms compliance at a 1 Hz cycle rate instead of 20 Hz (see [32]).

For the robot's perceived ability to follow instructions, touch modality received the highest rating. The combined voice+gesture modalities received the lowest. This could be due to the robot only performing the instructed action after the voice command had completed, while the action after the gesture mode interaction was followed immediately. However, they were not statistically significant at the 5 % level, and only indicated a trend towards higher mean preference to the touch modality.

In general, without considering the task, the results are in contrast to the result in [25], [26], and [27]. However, this contrast indicates an agreement with [22] and [24], namely that for certain tasks humans can communicate to robots effectively using a uni-modal communication channel.

# 7 FUTURE WORK

This research is eventually aiming to evaluate how best to teach a robot and what constitutes an effective teaching strategy. The work presented here is an initial attempt towards that direction, and further research is required. The software system could be further developed to accommodate more complex input interfaces. It would also be useful to conduct the same experiment with different user groups, e.g. children or people with special needs.